\documentclass[runningheads]{llncs}

\usepackage[T1]{fontenc}
\usepackage{amssymb,amsmath,amsfonts}
\usepackage{hyperref}
\usepackage{multirow}
\usepackage{graphicx,verbatim}
\usepackage{color}
\usepackage{booktabs}

\begin{document}

\title{Topology-Aware Training and Spatial Diagnostics for Fiber Bundle Segmentation in Tracer Histology}

\titlerunning{Topology-Aware Fiber Bundle Segmentation}

\author{
Joselyn Romero Avila\inst{1} \and
Kyriaki-Margarita Bintsi\inst{2} \and
Ermias Habte\inst{2} \and
Julia F.~Lehman\inst{3} \and
Suzanne N.~Haber\inst{3,4} \and
Anastasia Yendiki\inst{2}
}

\authorrunning{J. Romero Avila et al.}

\institute{
Universidad Nacional Mayor de San Marcos, Lima, Peru\\
\email{joselyn.romero1@unmsm.edu.pe}
\and
Athinoula A. Martinos Center for Biomedical Imaging,
Massachusetts General Hospital and Harvard Medical School,
Charlestown, MA, United States
\and
Department of Pharmacology and Physiology,
University of Rochester School of Medicine,
Rochester, NY, United States
\and
McLean Hospital, Belmont, MA, United States
}



\maketitle

\begin{abstract}
Anatomic tracer studies reveal how axon bundles project from an injection site, branch into smaller groups of axons, and course through the brain to reach their destinations. Histological data from such studies provide anatomical reference information for validating diffusion MRI tractography. However, manual annotation of the histological data is very labor-intensive, and although automated segmentation methods have been proposed, they rely mainly on pixel-overlap losses such as BCE and Dice; topology-aware loss functions have not been studied for this task. We compare BCE--Dice, clDice, Betti matching, and Topograph for fiber bundle segmentation in macaque tracer histology using a frozen DINOv3 backbone. To our knowledge, this is the first exploration of foundation-model features for this task. BCE--Dice achieved the highest Dice, while clDice achieved the highest bundle recall but poor mask overlap. Topograph had similar Dice to BCE--Dice, the lowest $B_0$ error, and fewer false positives than BCE--Dice and Betti matching. Fiber bundle segmentation methods are typically evaluated with a permissive rule that counts a bundle as detected given any overlap with the prediction. We show this rule does not capture oversegmentation, and that per-section TPR can be inflated by empty sections assigned perfect recall. To quantify this, we introduce $\mathrm{Excess}_{32}$,  a spatial diagnostic measuring predicted pixels outside a 32-pixel tolerance band around annotated bundles. In validation, a Betti--Topograph union raises sparse-bundle TPR from 0.818 to 0.933, but worsens FDR from 0.296 to 0.509, $\mathrm{Excess}_{32}$ from 0.108 to 0.466, and area ratio from 0.94 to 3.34. These results show detection metrics alone are insufficient to characterize segmentation quality.
\end{abstract}

\keywords{Anatomic tracing \and
fiber bundle segmentation \and topology-aware learning \and
segmentation metrics}

\section{Introduction}
\label{sec:intro}

Diffusion MRI (dMRI) tractography allows white-matter pathways to be reconstructed non-invasively and in vivo. However, it relies on indirect measurements of axonal orientations based on water diffusion, at a spatial scale much coarser than that of individual axons. This creates a need for validation against ground-truth connectional anatomy. Tracer injection studies in non-human primates provide anatomical reference data for validating dMRI tractography, as they allow fiber bundles projecting from an injection site to be visualized at microscopic resolution
\cite{yendiki2022post,grisot2021diffusion,jbabdi2013human,thomas2014anatomical,schilling2019limits}. This validation, in turn, requires the segmentation of fiber bundles in large histological sections. The process is labor-intensive and requires neuroanatomic expertise to distinguish fibers from other signals in the images, such as terminal fields, artifacts, and background staining
\cite{sundaresan2025self,bintsi2025fully}. Recent work has sought to automate this task using semi-supervised learning, self-supervised pretraining, synthetic data, and anatomical post-processing across sections
\cite{sundaresan2022constrained,sundaresan2025self,bintsi2025fully,bintsi2026tractography}. Despite these advances, fiber bundle segmentation remains challenging because manually annotated data are limited and substantial variability exists across brains, tracers, injection areas, and fiber bundles of different densities.

Foundation models are of particular interest in settings where annotated data are limited, as self-supervised pretraining can provide transferable visual representations. Models in the DINO family have shown strong transfer to dense prediction tasks \cite{caron2021dino,oquab2023dinov2,simeoni2025dinov3}. More recently, DINOv3 extended this family through training on extensive collections of web images \cite{simeoni2025dinov3}. Its use for fiber bundle segmentation in tracer histology has not yet been explored. A further consideration is the training loss. Prior work on tracer histology segmentation has focused mainly on pixel overlap terms such as binary cross-entropy (BCE) and Dice loss. These losses do not directly account for the structure of fiber bundles, which often appear as elongated and connected regions. Topology-aware losses offer an alternative way to incorporate structural information during training. For example, clDice measures skeleton overlap \cite{shit2021cldice}, Betti matching penalizes errors in connected components and holes \cite{stucki2024betti}, and Topograph compares graph-based representations \cite{lux2024topograph}. These approaches have been used to preserve structural or topological properties in curvilinear segmentation tasks, including retinal vessels, road networks, and neuronal processes \cite{shit2021cldice,stucki2024betti,lux2024topograph}. The effect of these loss functions on fiber bundle segmentation in tracer histology remains unclear.

Furthermore, standard metrics used to evaluate segmentation models rely on a permissive many-to-many rule, where a fiber bundle included in the ground-truth labels is considered detected if it has any overlap with one of the predicted bundles
\cite{sundaresan2025self,bintsi2025fully}. This rule is useful for sparse
and fragmented annotations, but it does not penalize predicted bundles that extend far outside the ground-truth annotations. A large, contiguous area in the prediction can increase the true-positive rate (TPR) without being counted as a false-positive component, even though it may include large swaths of tissue that do not overlap with the ground-truth annotation. This is consistent with recent work showing that aggregate scores can hide domain-relevant errors and that metrics should match the structure, task, and scientific question being evaluated
\cite{maierhein2024metrics,reinke2024pitfalls,bernhard2024outside}.


In this work, we study loss functions and evaluation metrics for fiber-bundle segmentation in tracer histology. We use a frozen DINOv3 backbone \cite{simeoni2025dinov3}, and compare multiple loss functions (BCE--Dice, clDice, Betti matching, Topograph) under the same architecture. We then evaluate how well pixel-overlap vs.~component-level error metrics capture the quality of the resulting predictions.
This paper makes the following contributions. 
\textbf{(1)} We provide the first systematic comparison of topology-aware losses for fiber bundle segmentation in macaque tracer histology using frozen foundation-model features.
\textbf{(2)} We audit standard evaluation metrics and identify two failure modes: oversegmentation hidden by component-level matching, and per-section TPR inflation caused by sections with no annotated bundles.
\textbf{(3)} We introduce $\mathrm{Excess}_{32}$, a spatial diagnostic that measures predicted pixels outside a 32-pixel tolerance band around annotated bundles and reveals oversegmentation missed by legacy metrics. The code is publicly available at
\url{https://github.com/jromero158486/topology-aware-fiber-bundle-segmentation}.

\section{Methods}
\label{sec:methods}

\subsubsection{Task Definition.}
\label{sec:task}
\begin{figure}[t]
\centering
\includegraphics[width=\textwidth]{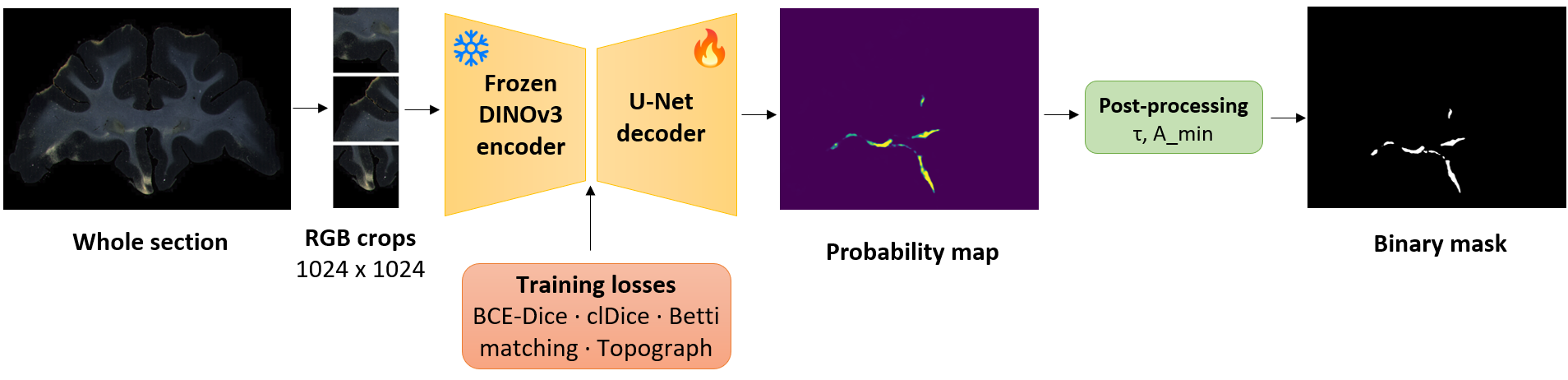}
\caption{Proposed pipeline. RGB patches are extracted from macaque tracer histology sections and processed by a frozen DINOv3 encoder and a trainable U-Net decoder. During training, patches are sampled from both annotated foreground and background or mixed regions, and we compare BCE--Dice, clDice, Betti matching, and Topograph losses. At inference, the probability maps from overlapping crops are averaged to reconstruct a section-level probability map. This map is thresholded, and connected components smaller than $A_{\min}$ are removed to obtain the final binary mask.}

\label{fig:pipeline}
\end{figure}

Let $I_i \in \mathbb{R}^{H_i \times W_i \times 3}$ denote an RGB coronal histological section from macaque tracer data. Each section has expert annotations for dense~(D), moderate~(M), and sparse~(S) fiber bundles. For training, we merge these labels into a single binary foreground mask $M_i^* \in \{0,1\}^{H_i \times W_i}$, since our objective is to detect whether a bundle is present rather than to classify its density. The model predicts a probability map $P_i \in [0,1]^{H_i \times W_i}$. At evaluation, detections are reported separately for dense, moderate, and sparse bundles to reveal whether segmentation performance depends on fiber bundle density. An overview of the proposed approach is shown in Figure~\ref{fig:pipeline}.

\subsubsection{Model Architecture.}
\label{sec:arch}
We use a frozen DINOv3 ViT-B/16 backbone~\cite{simeoni2025dinov3} as the
encoder and train a U-Net decoder~\cite{ronneberger2015unet} for binary segmentation. Freezing
the encoder keeps the number of trainable parameters small, which is beneficial when manually annotated data are limited. The probability map is obtained as
$P_i = d_\theta\!\left(e_{\mathrm{DINOv3}}(I_i)\right)$,
where $e_{\mathrm{DINOv3}}$ is the frozen encoder and $d_\theta$ is the
trainable decoder.

We extract intermediate feature maps from four encoder layers and project each to 256 channels via $1{\times}1$ convolutions. These patch-level features are decoded through a U-Net-style decoder with skip connections and bilinear upsampling. A shallow convolutional pyramid built from the input RGB image is fused with the decoder features to recover the input patch resolution.



\subsubsection{Training Losses.}
\label{sec:losses}
We train four loss configurations using the same architecture. All configurations use BCE--Dice as the base segmentation loss, defined as the equally weighted combination of binary cross-entropy and Dice loss. The \emph{BCE--Dice} configuration uses this base loss alone, while the remaining configurations add one auxiliary structural term: \emph{clDice}~\cite{shit2021cldice} adds a skeleton overlap term, \emph{Betti matching}~\cite{stucki2024betti} adds a topological term that penalizes errors in connected components and holes, and \emph{Topograph}~\cite{lux2024topograph} adds a graph-based loss that penalizes topological disagreement between the predicted and ground-truth masks.

\subsubsection{Inference.}
\label{sec:inference}
At inference, overlapping $1024{\times}1024$ patches are extracted with a stride equal to 25\% of the patch width. Their probability maps are averaged pixelwise to obtain a section-level prediction. The averaged map is thresholded at $\tau$, and 8-connected components smaller than $A_{\min}$ pixels are removed. The final values of $\tau$ and $A_{\min}$ are selected on the development set and fixed before held-out evaluation, as described in Section~\ref{sec:impl}.


\subsubsection{Evaluation Metrics.}
\label{sec:metrics}
All predictions and ground-truth masks are restricted to the annotated brain outline before evaluation, and connected components are computed using 8-connectivity throughout. We group the evaluation metrics into four categories, namely component detection, overlap, topology, and oversegmentation.


\noindent\textbf{\textit{Component detection.}}
Following prior tracer segmentation work \cite{sundaresan2025self,bintsi2025fully}, an annotated bundle is considered detected if it has any nonzero overlap with the binary prediction under many-to-many matching. A single predicted component may match more than one annotated bundle, and one annotated bundle may be matched by multiple predicted components. A predicted component is counted as a true positive (TP) if it overlaps with at least one annotated bundle and as a false positive (FP) otherwise.

We report true positive rate (TPR) in two forms. \emph{Macro TPR} is computed per section and then averaged across sections. Following prior work, sections with no annotated bundle of a given density receive TPR~1~\cite{bintsi2025fully,sundaresan2025self}. This convention can inflate the reported average when empty sections are common, as examined in Section~\ref{sec:metricfailure}. \emph{Pooled TPR} divides the total number of detected annotated bundles by the total number of annotated bundles across all sections and is unaffected by empty sections.

We similarly report false discovery rate (FDR) in two forms.
\emph{Macro FDR} averages the per-section fraction of predicted
components classified as FP. \emph{Pooled FDR} divides the total number
of FP components by the total number of predicted components across all
sections. Pooled FDR is used for operating-point selection
(Section~\ref{sec:impl}), while macro FDR is retained for comparison
with prior work. We also report the average numbers of TP and FP
components per section (TPavg and FPavg). TPavg can exceed the number of
annotated bundles when multiple predicted components match the same
bundle.

\noindent\textbf{\textit{Overlap.}}
Pixel overlap is evaluated using Dice on the postprocessed prediction and ground truth for each section, followed by averaging across sections. Centerline agreement is evaluated using clDice~\cite{shit2021cldice}, which measures consistency of each mask with the skeleton of the other.

\noindent\textbf{\textit{Topology.}}
We report $B_0$ error as the mean absolute difference between the number of foreground connected components in the prediction and in the merged ground-truth fiber mask, computed per section and averaged across sections. This measures whether the prediction has approximately the right number of components
overall; it does not require or verify that predicted and annotated components correspond spatially to one another.

\noindent\textbf{\textit{Oversegmentation.}}
The metrics above do not explicitly measure how far a prediction extends beyond a
ground-truth bundle: a single large connected prediction can satisfy
component detection while covering substantial tissue with no ground-truth
bundle. We quantify this directly. For section $i$, let $\hat{M}_i$ be
the predicted mask, $M_i^*$ the ground-truth mask, and $B_i$ the annotated
brain outline. We dilate $M_i^*$ by $r$ pixels to obtain a tolerance
region, restrict it to $B_i$, and measure the fraction of predicted
foreground pixels that fall outside this region:
\begin{equation}
\mathrm{Excess}_{r,i} =
\frac{
\left|
\hat{M}_i \setminus
\left(
\operatorname{dilate}_{r}(M_i^*) \cap B_i
\right)
\right|
}{
\max(|\hat{M}_i|,1)
}.
\label{eq:excess}
\end{equation}
That is, $\mathrm{Excess}_{r,i}$ is the proportion of a section's predicted foreground that lies outside an $r$-pixel tolerance band around the annotated bundles; a value near 0 indicates predictions are tightly localized to annotated regions, while a value near 1 indicates most of the prediction lies outside them. The metric is computed per section and then averaged across sections. 


We use $r=32$ pixels, approximately 0.225 mm, as the primary tolerance. This value was chosen empirically to allow modest differences around annotated bundle boundaries and was not derived from a specific anatomical scale. We additionally report $r=16$ and $r=64$ to assess whether the observed pattern is sensitive to the tolerance radius. Finally, we report the pooled predicted-to-ground-truth foreground area ratio as a complementary, tolerance-free measure of overall over- or under-segmentation.

\begin{equation}
\mathrm{AreaRatio}_{\mathrm{pooled}} =
\frac{
\sum_i |\hat{M}_i|
}{
\max\left(\sum_i |M_i^*|,1\right)
}.
\label{eq:arearatio}
\end{equation}
Pooling foreground areas before division prevents sections with empty or very small ground-truth masks from disproportionately affecting the aggregate ratio.

\section{Experiments}
\label{sec:experiments}

\subsection{Data}
\label{sec:data}


We use coronal histological sections from five macaque brains (M1--M5), with about 30 sections per brain and 165 annotated sections in total.
These brains are part of a larger dataset from 20 animals with bidirectional tracer injections at multiple cortical sites. Each section is 50\,\textmu m thick and was digitized at an in-plane resolution of 0.44\,\textmu m and converted to the multiscale format OME-Zarr. Model inputs were read from resolution level~4, corresponding to 7.035\,\textmu m per pixel. Relative to the native resolution, this represents a $16{\times}$ reduction in each spatial dimension and a $256{\times}$ reduction in pixel count. Expert neuroanatomists outlined areas that contained groups of axons traveling in close proximity to each other, and categorized them as dense~(D), moderate~(M), or sparse~(S) fiber bundles \cite{lehman2011rules,haynes2013organization}.

For model selection, loss comparison, and operating-point selection, we use a development set comprising all annotated sections from M1, the anterior portion of M2 (sections 17--27), and the posterior portion of M3 (sections 1--16), following prior work~\cite{bintsi2025fully}. Five-fold validation is performed within this development set. For evaluation, we use the held-out posterior portion of M2 (sections 1--15) as a within-animal split, and M4 (sections 1--33) and M5 (sections 1--42) as cross-subject brains. M2 section 16 and M5 section 43 were excluded because of image artifacts.

\subsection{Baselines}
\label{sec:baselines}
We compare the final model with prior macaque tracer segmentation methods~\cite{bintsi2025fully,sundaresan2025self}. Component-level and overlap metrics, as well as the topological metric $B_0$, are evaluated under the same protocol. Oversegmentation metrics are reported only for predictions available within our pipeline.

Comparisons with published methods provide context for held-out performance. The main controlled analysis focuses on loss behavior, operating-point selection, and metric reliability within our framework.

\subsection{Implementation Details}
\label{sec:impl}

\noindent\textbf{Patch sampling and augmentation.}
We train on $1024{\times}1024$ crops, sampling 40 patches per section.
Half are centered on annotated foreground pixels, and the remainder are
sampled from background or mixed regions~\cite{bintsi2025fully}. Images
are normalized using ImageNet statistics~\cite{deng2009imagenet}.
Augmentation includes horizontal and vertical flips and rotations by
multiples of $90^\circ$.

\noindent\textbf{Optimization.}
The decoder is optimized with AdamW~\cite{loshchilov2019decoupled} using
a learning rate and weight decay of $10^{-4}$, cosine annealing to 5\%
of the initial learning rate, mixed precision, and batch size~2. The checkpoint with the highest clDice on the validation sections is retained. This criterion emphasizes centerline agreement in
elongated bundles over area overlap alone and is applied to all loss configurations. Models used for loss comparison are trained for up to 150 epochs with early stopping patience~40. 

\noindent\textbf{Loss hyperparameters.}
The auxiliary loss weights were chosen empirically and are 0.1 for clDice, 0.5 for Betti matching,
and 0.1 for Topograph.

\noindent\textbf{Operating-point selection and loss comparison.}
For each topology-aware loss, we evaluated the same grid of probability thresholds $\tau$ and minimum component sizes $A_{\min}$ on the development predictions. The operating points were first examined under the legacy macro-averaged component protocol. Because favorable component scores did not always correspond to accurate mask-level agreement, the candidate losses were compared jointly using TPR, FDR, FPavg, Dice, and $B_0$ error. This initial screening was used to select the loss advanced to the five-fold out-of-fold analysis.

Topograph was subsequently evaluated, and its final operating point was selected from the combined out-of-fold predictions using the pooled recall requirements. For held-out evaluation, the five
probability maps were averaged before applying the selected threshold and minimum component size.

\section{Results}
\label{sec:results}

\subsection{Validation Loss Screening}
\label{sec:lossselection}


Table~\ref{tab:loss-selection} compares the four training losses at their selected development operating points using the metrics defined in Section~\ref{sec:metrics}. TPR values are averaged across sections. Dice and $B_0$ error complement TPR by measuring pixel overlap and agreement in the number of connected components.

\begin{table}[t]
\centering
\caption{Development screening of the candidate training losses. Each row
shows the selected operating point from the common threshold and
minimum-component-size grid.}
\label{tab:loss-selection}
\scriptsize
\setlength{\tabcolsep}{3.2pt}
\begin{tabular*}{\textwidth}
{@{\extracolsep{\fill}}lcrrrrrrr@{}}
\toprule
Loss
& $\tau$
& $A_{\min}$
& \shortstack{Mean\\TPR $\uparrow$}
& \shortstack{Sparse\\TPR $\uparrow$}
& \shortstack{FDR\\$\downarrow$}
& \shortstack{FPavg\\$\downarrow$}
& \shortstack{Dice\\$\uparrow$}
& \shortstack{$B_0$ err.\\$\downarrow$} \\
\midrule
BCE--Dice
& 0.10 & 50
& 0.876 & 0.850
& 0.464 & 7.01
& 0.599 & 4.425 \\
clDice
& 0.05 & 2000
& 1.000 & 1.000
& 0.194 & 0.44
& 0.036 & 7.408 \\
Betti matching
& 0.40 & 50
& 0.854 & 0.777
& 0.206 & 2.57
& 0.399 & 3.569 \\
Topograph
& 0.05 & 700
& 0.851 & 0.775
& 0.202 & 1.64
& 0.582 & 3.288 \\
\bottomrule
\end{tabular*}
\end{table}


No single loss performed best across all evaluation measures. clDice produced perfect macro-averaged component recall and the lowest FDR and FPavg, but its Dice was only 0.036 and its $B_0$ error was 7.408. In contrast, BCE--Dice had the highest Dice (0.599), with higher FDR and FPavg than Topograph. Topograph retained Dice close to BCE--Dice (0.582), achieved the lowest $B_0$ error (3.288), and had lower FDR and FPavg than BCE--Dice and Betti matching. Considering these trade-offs, Topograph was selected for the five-fold out-of-fold analysis.

\noindent\textbf{Topograph operating point.}
We selected the Topograph operating point by inspecting the trade-off between pooled component recall and pooled false-discovery rate across the predefined $(\tau,A_{\min})$ grid. Because higher-recall points also introduced a large false-positive component burden, we selected an application operating point that retained the highest mean pooled TPR
while keeping pooled FDR below 0.50.
The selected configuration was $\tau^*{=}0.20$ and $A_{\min}^*{=}2000$, with mean pooled TPR 0.775, sparse pooled TPR 0.633, and pooled FDR 0.478.


\begin{table}[t]
\centering
\caption{Evaluation on held-out brains. TPR values are averaged per
section under many-to-many component matching; D/M/S denote dense,
moderate, and sparse bundles. M2 is the within-animal split, while M4
and M5 are cross-subject brains.}
\label{tab:main}
\scriptsize
\setlength{\tabcolsep}{3.2pt}
\begin{tabular*}{\textwidth}{@{\extracolsep{\fill}}llcccrrrrr@{}}
\toprule
Brain & Method
& \multicolumn{3}{c}{\shortstack{TPR\\$\uparrow$}}
& TPavg
& \shortstack{FPavg\\$\downarrow$}
& \shortstack{FDR\\$\downarrow$}
& \shortstack{Dice\\$\uparrow$}
& \shortstack{$B_0$ err.\\$\downarrow$} \\
\cmidrule(lr){3-5}
& & D & M & S & & & & & \\
\midrule

\multirow{4}{*}{M2}
& Bintsi et al.~\cite{bintsi2025fully}
  & 0.98 & 0.95 & 0.63 & 4.25 & 1.19 & 0.110 & 0.310 & 3.33 \\
& \quad + pretrain
  & 0.97 & 0.95 & 0.61 & 5.56 & 2.56 & 0.180 & 0.280 & 4.00 \\
& Sundaresan et al.~\cite{sundaresan2025self}
  & 0.93 & 0.76 & 0.27 & 0.89 & 3.63 & 0.790 & 0.201 & 15.87 \\
& \textbf{Proposed}
  & 0.95 & 0.98 & 0.62 & 6.60 & 1.00 & 0.064 & 0.241 & 5.60 \\

\midrule

\multirow{4}{*}{M4}
& Bintsi et al.~\cite{bintsi2025fully}
  & 0.97 & 0.90 & 0.78 & 3.00 & 1.36 & 0.300 & 0.550 & 3.33 \\
& \quad + pretrain
  & 0.97 & 0.83 & 0.62 & 3.10 & 0.64 & 0.200 & 0.490 & 3.00 \\
& Sundaresan et al.~\cite{sundaresan2025self}
  & 0.85 & 0.75 & 0.41 & 0.85 & 2.91 & 0.620 & 0.265 & 5.52 \\
& \textbf{Proposed}
  & 0.97 & 0.91 & 0.75 & 4.18 & 3.06 & 0.366 & 0.436 & 3.67 \\

\midrule

\multirow{4}{*}{M5}
& Bintsi et al.~\cite{bintsi2025fully}
  & 0.99 & 0.99 & 0.86 & 0.86 & 0.81 & 0.350 & 0.279 & 1.73 \\
& \quad + pretrain
  & 0.99 & 1.00 & 0.87 & 0.95 & 1.05 & 0.250 & 0.218 & 1.67 \\
& Sundaresan et al.~\cite{sundaresan2025self}
  & 0.99 & 0.91 & 0.84 & 0.49 & 1.49 & 0.470 & 0.142 & 4.14 \\
& \textbf{Proposed}
  & 1.00 & 0.99 & 0.94 & 1.21 & 4.81 & 0.523 & 0.230 & 4.81 \\

\bottomrule
\end{tabular*}
\end{table}

Dense- and moderate-bundle recovery remained high across all three held-out brains. On the within-animal M2 split, the proposed method achieved the lowest FDR (0.064) and FPavg (1.00), with TPR for dense, moderate, and sparse fiber bundles comparable to those of the retrained baselines. On the cross-animal M4 brain, FDR and FPavg increased to 0.366 and 3.06. The per-section moderate-bundle TPR was 0.91, compared with a pooled TPR of 0.805; this difference is examined in Section~\ref{sec:metricfailure}. M5 showed the largest false-positive burden, with FDR 0.523 and FPavg 4.81, despite high per-section TPR.

\subsection{Metric Failure Modes}
\label{sec:metricfailure}

The development analyses revealed two limitations of the legacy
component-level evaluation: a prediction can achieve high bundle recall
despite substantial oversegmentation, and macro-averaged TPR can be
inflated by sections without annotated bundles.

\begin{figure}[t]
\centering
\includegraphics[
    width=\textwidth,
    trim=5 0 5 0,
    clip
]{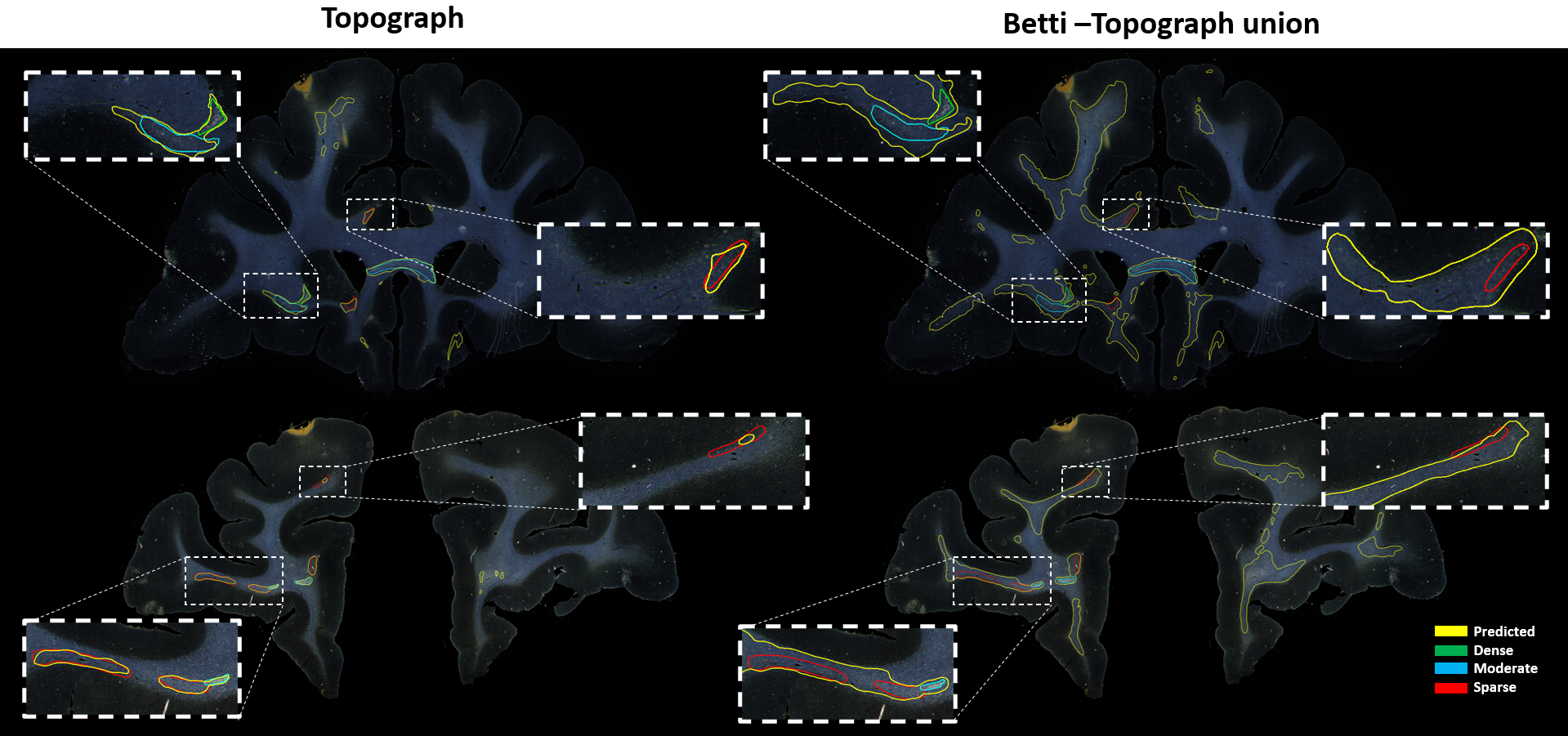}
\vspace{-0.3em}
\caption{Qualitative comparison of Topograph and the Betti--Topograph
union on the development set. Topograph predictions are shown on the
left and Betti--Topograph union predictions on the right for the same
histological sections. Dashed boxes indicate regions shown in enlarged
views. These views highlight the broader predictions produced by the
union, which extend into unannotated tissue. Predictions are outlined
in yellow; ground-truth bundles are dense (green), moderate (cyan), and
sparse (red).}
\label{fig:spatialoverextension}
\vspace{-0.8em}
\end{figure}

\noindent\textbf{Oversegmentation.}
During development, we evaluated postprocessing variants intended to
increase bundle recall while controlling false positives and
oversegmentation. These included pixelwise union and intersection with
Betti predictions, support gates that retained components corroborated
by another prediction, and continuity filters based on adjacent
sections. The Betti--Topograph union provided the clearest example of an
apparent improvement in component recall that did not correspond to
improved segmentation quality. Representative examples are shown in
Fig.~\ref{fig:spatialoverextension}.

\begin{table}[t]
\centering
\caption{Postprocessing analysis on the development set. All metrics are
computed from the same predictions and the same set of evaluated
sections. TPR values are macro-averaged across sections under the legacy
component protocol. $\mathrm{Excess}_{32}$ is the primary spatial
diagnostic, while $r=16$ and $r=64$ assess sensitivity to the tolerance
radius. The area ratio is computed after pooling foreground areas across
sections.}
\label{tab:spatialaudit}
\scriptsize
\setlength{\tabcolsep}{3.5pt}

\begin{tabular}{@{}lccccccccc@{}}
\toprule
Variant
& \multicolumn{3}{c}{TPR $\uparrow$}
& \shortstack{FDR\\$\downarrow$}
& \shortstack{FPavg\\$\downarrow$}
& \multicolumn{3}{c}{$\mathrm{Excess}_{r}$ $\downarrow$}
& \shortstack{Pooled area\\ratio $(\to 1)$} \\
\cmidrule(lr){2-4}
\cmidrule(lr){7-9}
& D & M & S & & & $r{=}16$ & $r{=}32$ & $r{=}64$ & \\
\midrule
Topograph
& 0.895 & 0.924 & 0.818
& 0.296 & 3.64
& 0.134 & 0.108 & 0.091
& 0.94 \\
Betti--Topograph union
& 0.937 & 0.957 & 0.933
& 0.509 & 18.57
& 0.510 & 0.466 & 0.421
& 3.34 \\
\bottomrule
\end{tabular}
\end{table}

As shown in Table~\ref{tab:spatialaudit}, the union increased TPR across
all density classes, including sparse-bundle TPR from 0.818 to 0.933.
However, FDR increased from 0.296 to 0.509 and FPavg from 3.64 to 18.57.
$\mathrm{Excess}_{32}$ also increased from 0.108 for direct Topograph to
0.466 for the Betti--Topograph union. The same pattern was observed at
both sensitivity radii: $\mathrm{Excess}_{16}$ increased from 0.134 to
0.510 and $\mathrm{Excess}_{64}$ from 0.091 to 0.421. The pooled
predicted-to-ground-truth area ratio increased from 0.94 to 3.34,
indicating that the union predicted more than three times the annotated
foreground area. The enlarged regions in Fig.~\ref{fig:spatialoverextension} illustrate this behavior, showing broader predictions beyond annotated bundles and additional false-positive components. Direct Topograph predictions were therefore retained for the final pipeline.

\noindent\textbf{Per-section TPR inflation.}
Under the legacy convention, a section without an annotated bundle of a given density receives TPR~1 for that class. On M4, 8 of 33 sections (24.2\%) contained no annotated moderate bundles and received moderate-bundle TPR~1. This contributed to the difference between macro moderate-bundle TPR (0.907, reported as 0.91 in Table~\ref{tab:main}) and pooled moderate-bundle TPR (0.805; 70 of 87 annotated bundles detected). The pooled value is computed from detected and annotated component counts across all sections and is not affected by empty sections.

\section{Discussion and Conclusion}
\label{sec:discussion}

The development analysis suggests that the loss functions favored different aspects of segmentation quality. BCE--Dice was strongest for mask overlap, whereas clDice favored bundle detection and false-positive control. Topograph retained mask overlap close to BCE--Dice with lower component-count error and fewer false positives than BCE--Dice and Betti matching. This combination motivated its selection.

On the held-out brains, the frozen DINOv3 encoder with a trainable decoder achieved bundle detection comparable to prior methods. The clearest reduction in false-positive burden was observed on M2, whereas Dice and $B_0$ error did not improve consistently across brains. These results suggest that web-pretrained features can support tracer histology segmentation despite differences between the pretraining and target domains. Because the encoder remained fixed, the present findings are limited to loss functions evaluated with a common DINOv3 representation. Further work is needed to determine whether the same trends hold with other architectures or after adapting the encoder to tracer histology.

The evaluation audit showed that high bundle detection did not always correspond to better segmentation. The Betti--Topograph union increased sparse-bundle TPR but also produced larger masks that extended beyond the annotated regions, as reflected by $\mathrm{Excess}_{32}$ and the pooled area ratio. Macro-averaged TPR showed a different limitation, since sections with no annotated bundles receive perfect recall under the legacy evaluation protocol. High component recall can therefore coexist with substantial oversegmentation, and reporting it alone may overstate segmentation quality.

In summary, component detection alone provides an incomplete assessment of fiber bundle segmentation in tracer histology. The loss comparison revealed trade-offs between mask overlap and component-level measures, while the evaluation audit showed that high recall can coexist with substantial oversegmentation or be affected by sections with no annotated bundles. The similar pattern observed at 16, 32, and 64 pixels suggests that this finding is not dependent on a single tolerance value. These results support the use of spatial diagnostics alongside component and overlap measures when evaluating fiber bundle segmentation. Future work should extend $\mathrm{Excess}_r$ and the pooled area ratio to other tracer segmentation methods and examine whether differences in segmentation quality influence downstream dMRI tractography validation.

\begin{credits}
\subsubsection{\ackname} 
This work was supported by the Center for Large-scale Imaging of Neural Circuits (LINC), an NIH BRAIN Initiative Connectivity across Scales (CONNECTS) comprehensive center (UM1-NS132358). Additional support was provided by the National Institute of Mental Health (R01-MH045573, P50-MH106435) and the National Institute of Neurological Disorders and Stroke (R01-NS119911, R01-NS127353).

\subsubsection{\discintname}
The authors have no competing interests to declare that are
relevant to the content of this article. 
\end{credits}



\bibliographystyle{splncs04}
\bibliography{dinov3_references}

\end{document}